%% file: main.tex
\documentclass[10pt,twocolumn,letterpaper]{article}

\usepackage{cvpr}              

\input{sec/preamble}

\definecolor{cvprblue}{rgb}{0.21,0.49,0.74}
\usepackage[pagebackref,breaklinks,colorlinks,allcolors=cvprblue]{hyperref}
\usepackage{ifthen} 
\usepackage[accsupp]{axessibility}  

\usepackage[utf8]{inputenc} 
\usepackage{CJKutf8} 
\usepackage{mathrsfs}
\usepackage{amsmath}
\usepackage{bm}
\usepackage{multirow}
\usepackage{bbding}
\usepackage{makecell}

\hypersetup{
colorlinks,
linkcolor={red},
urlcolor={purple}
}

\usepackage[capitalize]{cleveref}
\crefname{section}{Sec.}{Secs.}
\Crefname{section}{Section}{Sections}
\Crefname{table}{Table}{Tables}
\crefname{table}{Tab.}{Tabs.}
\Crefname{figure}{Figure}{Figures}
\crefname{figure}{Fig.}{Figs.}
\Crefname{equation}{Equation}{Equations}
\crefname{equation}{Eq.}{Eqs.}

\definecolor{rred}{RGB}{245, 152, 153}
\definecolor{oorange}{RGB}{253, 205, 154}
\definecolor{carolinablue}{rgb}{0.6, 0.73, 0.89}
\newcommand{\algname}{SeedVR}

\definecolor{cvprblue}{rgb}{0.21,0.49,0.74}
\usepackage{xspace}

\title{SeedVR: Seeding Infinity in Diffusion Transformer \\ Towards Generic Video Restoration}

\author{Jianyi Wang$^{1, 2*}$\blfootnote{$\dag$~corresponding authors}~~~~Zhijie Lin$^{2}$~~~~Meng Wei$^2$~~~~Yang Zhao$^{2}$~~~~Ceyuan Yang$^2$~~~~Fei Xiao$^2$ \\ Chen Change Loy$^1$~~~~Lu Jiang$^{2}$\vspace{2mm}\\
$^1$Nanyang Technological University\hspace{1.5cm}$^2$ByteDance\hspace{1.5cm}\\
{\tt\small \url{https://iceclear.github.io/projects/seedvr/}}
}

\begin{document}

\newboolean{putfigfirst}
\setboolean{putfigfirst}{true}
\ifthenelse{\boolean{putfigfirst}}{
\twocolumn[{%
\renewcommand\twocolumn[1][]{#1}%
\maketitle\thispagestyle{empty}
\vspace{-1cm}
\begin{center}
\includegraphics[width=\linewidth]{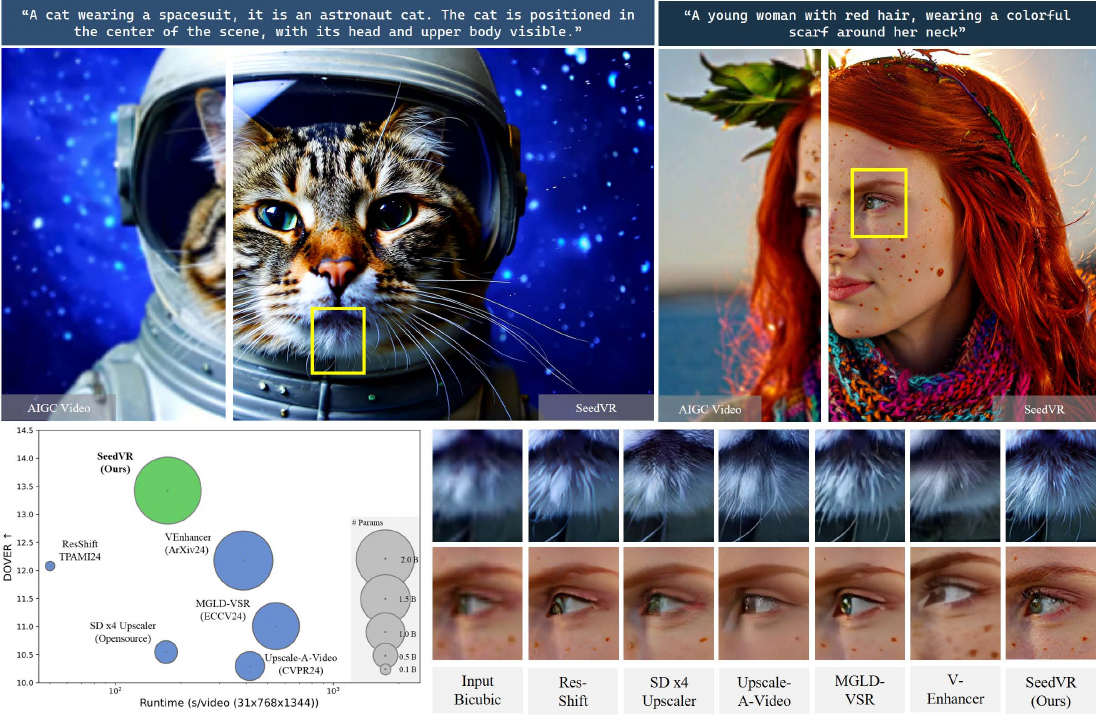}
\vspace{-7mm}
\captionof{figure}{Speed and performance comparisons. \algname~demonstrates impressive restoration capabilities, offering fine details and enhanced visual realism.
Despite its 2.48B parameters, SeedVR is over $2 \times$ faster than existing diffusion-based video restoration approaches~\cite{zhou2024upscaleavideo,yang2023mgldvsr,he2024venhancer}.
With delicate designs, SeedVR is as efficient as the Stable Diffusion Upscaler~\cite{sdupscaler}, even with five times the parameter count. \textbf{(Zoom-in for best view)}
}
\vspace{0.5mm}
\label{fig:teaser}
\end{center}%
}]
}

\maketitle
\input{sec/0_abstract}    
\input{sec/1_intro}
\input{sec/2_related_work}

\input{sec/3_method}
\input{sec/4_exp}
\input{sec/5_conclusion}

\clearpage

\noindent{\textbf{Acknowledgement:}} This research is supported by the National Research Foundation, Singapore under its AI Singapore Programme (AISG Award No: AISG2-PhD-2022-01-033[T]).
We sincerely thank Zhibei Ma for data processing, Jiashi Li for hardware maintenance.
We also give thanks to Shangchen Zhou and Zongsheng Yue for research discussions.

{
    \small
    \bibliographystyle{ieeenat_fullname}
    \bibliography{main}
}


\end{document}

%% file: sec/preamble.tex
\usepackage[accsupp]{axessibility}
\usepackage{microtype}
\usepackage[dvipsnames]{xcolor}
\usepackage{pifont}
\usepackage{multicol}
\usepackage{multirow}
\usepackage{caption}
\usepackage{paralist}

\usepackage{lipsum}

\newcommand\blfootnote[1]{%
  \begingroup
  \renewcommand\thefootnote{}\footnote{#1}%
  \addtocounter{footnote}{-1}%
  \endgroup
}

\definecolor{onlinecolor}{HTML}{D0C2B6}
\definecolor{frozencolor}{HTML}{8693AB}
\definecolor{mypink}{HTML}{FF968D}

%% file: sec/0_abstract.tex
\begin{abstract}
\blfootnote{$^*$ Work was done during Jianyi Wang's internship at ByteDance in Singapore. (iceclearwjy@gmail.com)} Video restoration poses non-trivial challenges in maintaining fidelity while recovering temporally consistent details from unknown degradations in the wild.
Despite recent advances in diffusion-based restoration, these methods often face limitations in generation capability and sampling efficiency.
In this work, we present \text{SeedVR}, a diffusion transformer designed to handle real-world video restoration with arbitrary length and resolution.
The core design of SeedVR lies in the shifted window attention that facilitates effective restoration on long video sequences.
SeedVR further supports variable-sized windows near the boundary of both spatial and temporal dimensions, overcoming the resolution constraints of traditional window attention.
Equipped with contemporary practices, including causal video autoencoder, mixed image and video training, and progressive training, SeedVR achieves highly-competitive performance on both synthetic and real-world benchmarks, as well as AI-generated videos.
Extensive experiments demonstrate SeedVR's superiority over existing methods for generic video restoration. 
\end{abstract}

%% file: sec/1_intro.tex
\section{Introduction}
\label{sec:intro}

Generic video restoration (VR) is a classical computer vision task, seeking to reconstruct high-quality (HQ) outputs from low-quality (LQ) input videos.
A broad range of works have been proposed to tackle the challenges posed by complex and often unknown degradations~\cite{zhou2022codeformer,wang2021realesrgan,chan2022investigating,xie2023mitigating} encountered in real-world VR scenarios~\cite{wang2019edvr,chan2021basicvsr,chan2022basicvsr++,liang2024vrt,liang2022recurrent,yang2021real}.

More recently, diffusion-based image~\cite{wang2024exploiting,yu2024scaling,wu2024seesr} and video restoration methods~\cite{yeh2024diffir2vr,yang2023mgldvsr,chen2024learning,zhou2024upscaleavideo,he2024venhancer}, often built on U-Net architectures with full-attention layers, have shown promise in addressing the issues found in previous approaches such as over-smoothing. However, the attention design in diffusion leads to significant computational costs and performance degradation when processing resolutions different from those used during training, limiting their applicability for restoring long-duration, high-resolution videos.

As such, previous VR approaches~\cite{yeh2024diffir2vr,yang2023mgldvsr,chen2024learning,zhou2024upscaleavideo,he2024venhancer} rely on patch-based sampling~\cite{jimenez2023mixture,wang2024exploiting}, \ie,  dividing the input video into overlapping spatial-temporal patches and fusing these patches using a Gaussian kernel at each diffusion step.
The large overlap (\eg, 50\% of the patch size), required for ensuring a coherent output without visible patch boundaries, often leads to considerably slow inference speed.
This inefficiency becomes even more pronounced when processing long videos at high resolutions.
For instance, VEnhancer~\cite{he2024venhancer} takes $387$ seconds to generate $31$ frames at a resolution of $1344 \times 768$ with 50 sampling steps, even when using only temporal overlap. 
Likewise, Upscale-A-Video~\cite{zhou2024upscaleavideo}, using a spatial overlapping of $384 \times 384$ and a temporal overlapping of $2$, takes $414$ seconds to process the same video clip,  rendering it less practical for real-world use.

In this work, we present \textbf{SeedVR}, a Diffusion Transformer (DiT) model designed for generic video restoration (VR) that tackles resolution constraints efficiently. We propose a design using \emph{large non-overlapping window} attention in DiT, which we found effective for achieving competitive VR quality at a lower computational cost. Specifically, SeedVR uses MMDiT~\cite{esser2024scaling} as its backbone and replaces full self-attention with a window attention mechanism. While various window attention designs have been explored, we aim to keep our design as simple as possible and hence use the Swin attention~\cite{liu2021swin}, resulting in Swin-MMDiT. Unlike previous methods~\cite{liang2021swinir,yue2023resshift,yue2024efficient}, our Swin-MMDiT adopts a significantly larger attention window of $64 \times 64$ over an $8 \times 8$ compressed latent, compared to the $8 \times 8$ pixel space commonly used in window attention for low-level vision tasks~\cite{liang2021swinir,yue2023resshift,yue2024efficient}. When processing arbitrary input resolutions with Swin-MMDiT using a large window, we can no longer assume that the input spatial dimensions will be multiples of the window size. Additionally, the shifted window mechanism in Swin results in uneven 3D windows near the boundaries of the space-time volume. To address these, we design a 3D rotary position embedding~\cite{su2024roformer} within each window to model the varying-sized windows.

To enhance the SeedVR training, we further incorporate several techniques inspired by recent work. First, building on Yu \etal~\cite{yu2024language}, we develop a causal video variational autoencoder (CVVAE) that compresses time and space by factors of $4$ and $8$, respectively. This CVVAE significantly reduces the computational cost of VR, especially for high-resolution videos, while maintaining high reconstruction quality. Second, motivated by Dehghani \etal~\cite{dehghani2024patch}, we train SeedVR on images and videos with native and varying resolutions. Finally, we employ a multi-stage progressive training strategy to accelerate convergence on large-scale datasets.
Extensive experimental results demonstrate that our SeedVR performs steadily well across various VSR benchmarks from different sources, serving as a strong baseline for VR in diverse real-world scenarios, as shown in \Cref{fig:teaser}.
\textbf{See the supplementary materials for the videos}.

To our knowledge, SeedVR is among the earliest explorations on training a large scalable diffusion transformer model designed for generic video restoration.
The main contributions are as follows:
1) We tackle the key challenge in diffusion-based VR, \ie, handling inputs with arbitrary resolutions, by proposing simple yet effective diffusion transformer blocks based on a shifted window attention mechanism. 
2) We further develop a casual video autoencoder, considerably improving both training and inference efficiency while achieving favorable video reconstruction quality.
As shown in \Cref{fig:teaser}, SeedVR is at least \textbf{$\bm{2 \times}$ faster} than existing diffusion-based VR methods~\cite{yang2023mgldvsr,zhou2024upscaleavideo,he2024venhancer}, despite having $2.48$B parameters, which is over $3.5\times$ more than Upscale-A-Video~\cite{zhou2024upscaleavideo}.
3) By leveraging large-scale joint training on images and videos, along with multi-scale progressive training, SeedVR achieves state-of-the-art performance across diverse benchmarks, outperforming existing approaches by a large margin.
Serving as the largest-ever diffusion transformer model towards generic VR, we believe SeedVR will push the frontiers of advanced VR and inspire future research in developing large vision models for real-world VR.

%% file: sec/2_related_work.tex
\section{Related Work}
\label{sec:related}
\noindent\textbf{Attention Mechanism in Restoration.} Early restoration approaches that adopted CNN-based architectures~\cite{wang2019edvr,chan2021basicvsr,chan2022basicvsr++,chan2022investigating,tian2020tdan,huang2015bidirectional,sajjadi2018frame,isobe2020video} typically struggled to capture long-range pixel dependencies due to limited receptive fields.
Recent advances in transformer models have inspired a series of restoration methods~\cite{chen2021pre,xie2023mitigating,liang2021swinir,chen2023activating,zhou2023srformer,zhou2024srformerv2,liang2024vrt} that introduce attention mechanisms into restoration networks, further improving performance on restoration benchmarks.
To mitigate the quadratic complexity of the self-attention mechanism~\cite{vaswani2017attention}, many of these approaches~\cite{liang2021swinir,chen2023activating,zhou2023srformer,zhou2024srformerv2,liang2024vrt} use window attention to reduce computational costs.
For instance, SwinIR~\cite{liang2021swinir} adopts Swin Transformer~\cite{liu2021swin} with a $8 \times 8$ window attention.
SRFormer~\cite{zhou2023srformer} and SRFormerV2~\cite{zhou2024srformerv2} further increase the window size to $24 \times 24$ and $40 \times 40$, respectively, to enhance performance.
Despite these improvements, limited window sizes still restrict the receptive field, especially in diffusion models where text embeddings interact with image embeddings within each window. 
As a result, existing diffusion-based restoration methods~\cite{rombach2021highresolution,wang2024exploiting,yu2024scaling,zhou2024upscaleavideo,he2024venhancer,wu2024seesr} continue to rely on full attention to achieve effective restoration with text guidance.

In this study, we focus on investigating the window attention mechanism within a diffusion transformer for VR.
By employing a substantially larger attention window, \ie, $64 \times 64$ in an $8 \times$ compressed latent space, our method interacts with text prompts and captures long-range dependencies.
We also introduce variable-sized windows near the boundaries of each dimension, reducing resolution constraints.
This design allows our approach to circumvent the reliance on tiled sampling strategies~\cite{jimenez2023mixture,wang2024exploiting} enables direct application to VR tasks with any length and resolution.

\noindent\textbf{Diffusion Transformer.}
The development of DiT~\cite{peebles2023scalable} has made diffusion transformer the prevailing architecture for diffusion models~\cite{esser2024scaling,liu2024sora,yang2024inf,hatamizadeh2025diffit,polyak2024movie,chen2023pixartalpha,li2024hunyuan,zhang2024tora,chen2023fit,kondratyukvideopoet,jin2024pyramidal,chen2024pixart}.
To reduce the high computational cost of generating high-resolution images and videos, common approaches include using separate temporal and spatial attention~\cite{zhang2024tora}, applying token compression~\cite{chen2024pixart} and generating outputs in a multi-stage manner~\cite{jin2024pyramidal}.
Instead of relying on the full attention mechanism, FIT~\cite{chen2023fit} interleaves window attention and global attention with two types of transformer.
While this method is efficient, it falls short of handling variable-sized inputs in VR.
Inf-DiT~\cite{yang2024inf} enables upscaling on images of varying shapes and resolutions by using local attention in an autoregressive manner, though it is limited by a finite receptive field. 
The approach most similar to ours is VideoPoet~\cite{kondratyukvideopoet}, which uses three types of 2D window attention for video super-resolution, each performing self-attention within a local window aligned along one of three axes.
However, this method still struggles with arbitrary input shapes, as it requires a full attention operation along one axis.
In contrast, our approach introduces a flexible 3D window attention that can be effectively applied to VR with varying resolutions.

\noindent\textbf{Video Restoration.}
%
Most previous works~\cite{chan2021basicvsr,chan2022basicvsr++,liang2024vrt,liang2022recurrent,li2023simple,chen2024learning,youk2024fma,wang2019edvr} focus primarily on synthetic data, resulting in limited effectiveness for real-world VR.
Later approaches~\cite{chan2022investigating,xie2023mitigating,zhang2024realviformer} have shifted towards real-world VR, yet still struggle to produce realistic textures due to limited generative capabilities.
Motivated by recent advances in diffusion models~\cite{sohl2015deep,ho2020denoising,2021Score,nichol2022glide,rombach2021highresolution}, several diffuison-based VR approaches~\cite{zhou2024upscaleavideo,he2024venhancer,yang2023mgldvsr} have emerged, showing impressive performance. While fine-tuning from a diffusion prior~\cite{rombach2021highresolution,wang2024videocomposer} provides efficiency, these methods still inherit limitations inherent to diffusion priors.
In particular, they use a basic autoencoder without temporal compression, resulting in inefficient training and inference. Additionally, their reliance on full attention imposes resolution constraints, further increasing the inference cost.
Unlike existing diffusion-based VR approaches, we redesign the whole architecture with an efficient video autoencoder and a flexible window attention mechanism, achieving effective and efficient VR with arbitrary length and resolution.

%% file: sec/3_method.tex
\section{Methodology}
\label{sec:method}

\begin{figure*}[th]
    \refstepcounter{subfigure}\label{fig:network:a}
    \refstepcounter{subfigure}\label{fig:network:b}
    \centering
    \includegraphics[width=\linewidth]{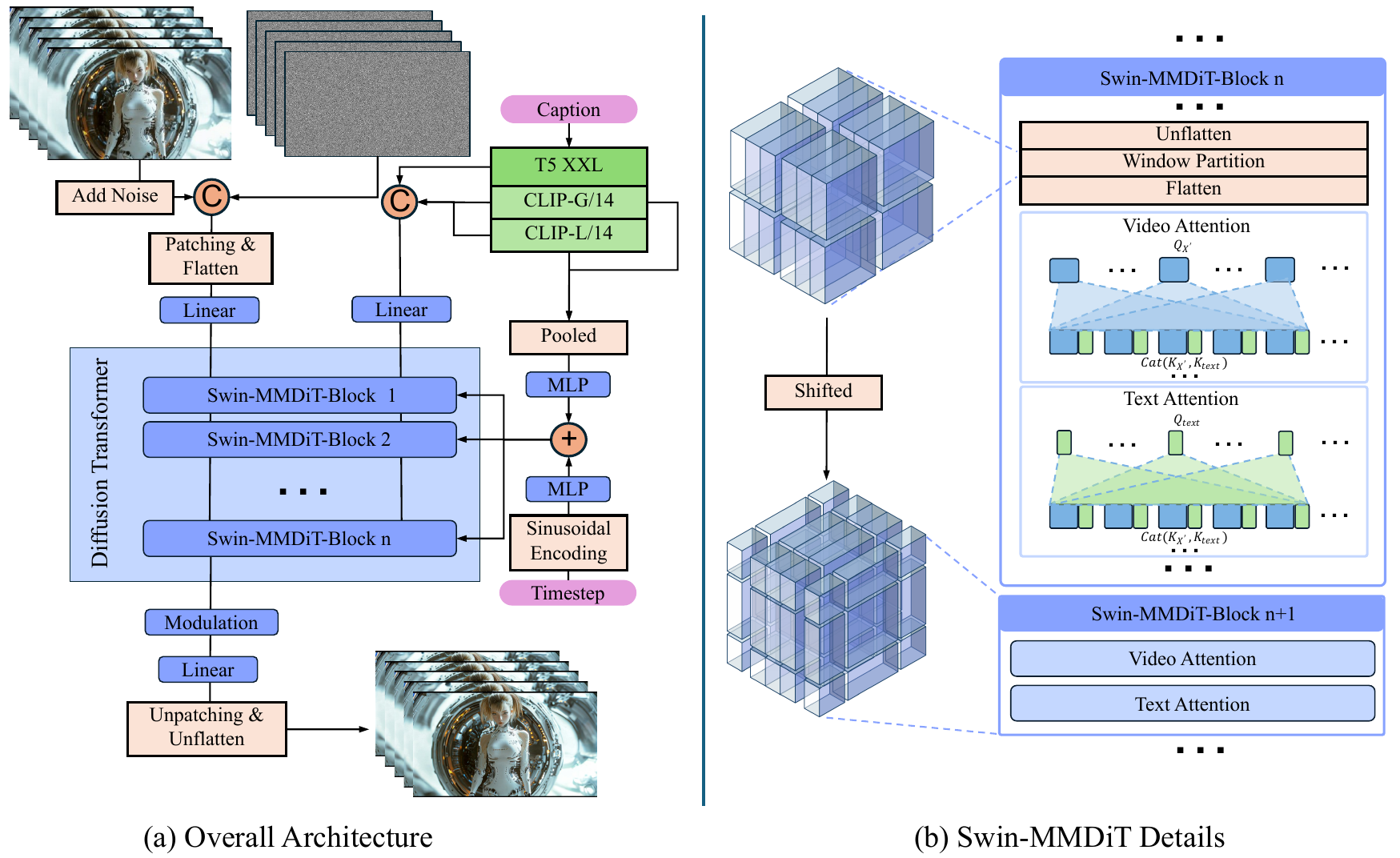}
    \vspace{-7mm}
    \captionof{figure}{
     Model architecture and the details of Swin-MMDiT of SeedVR. Our approach introduces a shifted window mechanism into the transformer block, bypassing the resolution constrain of vanilla attention. We further adopt large attention windows around the center and variable-sized windows near the boundary, enabling long-range dependency capturing given inputs of any length and size.
    } \vspace{-5mm}
    \label{fig:network}
\end{figure*}

We focus on effective VR with arbitrary lengths and resolutions, which is still underexplored.
As depicted in Figure~\ref{fig:network:a}, our approach employs a similar architecture following SD3~\cite{esser2024scaling}, where a pretrained autoencoder is applied to compress the input video into latent space, and the corresponding text prompt is encoded by three pretrained, frozen text encoders~\cite{radford2021learning,cherti2023reproducible,raffel2020exploring}.
To relax the resolution constraints of the MMDiT block used in SD3, we introduce a Swin-MMDiT block based on a shifted window mechanism, as detailed in Sec.~\ref{subsec:swin}.
We further present our casual video autoencoder in Sec.~\ref{subsec:vae}, which significantly improves the training and inference efficiency compared to existing approaches~\cite{zhou2024upscaleavideo,he2024venhancer,yang2023mgldvsr}.
In Sec.~\ref{subsec:training}, we discuss the training strategies to effectively train our model on large-scale datasets.

\subsection{Shifted Window Based MM-DiT}
\label{subsec:swin}
MMDiT has been proven to be an effective transformer block by SD3~\cite{esser2024scaling}, where separate weights are applied to the two modalities, \ie, visual input and text, thus enabling a bidirectional flow of information between visual features and text tokens.
However, the full attention nature of MMDiT makes it unsuitable for VR, which requires the capability of handling inputs with arbitrary lengths and resolutions.
To this end, we introduce a shifted window attention mechanism into MMDiT, which we call as Swin-MMDiT.
Given a video feature $X\in \mathbb{R}^{T\times H\times W\times d}$ and a text embedding $C_{text} \in \mathbb{R}^{L \times d}$, the video feature is first flattened to $X^{\prime} \in \mathbb{R}^{THW\times d}$, following the NaViT scheme~\cite{dehghani2024patch}.
For MMDiT~\cite{esser2024scaling}, it directly extracts $(Q_{X^{\prime}}, K_{X^{\prime}}, V_{X^{\prime}})$ and $(Q_{text}, K_{text}, V_{text})$ from $X^{\prime}$ and $C_{text}$, respectively.
Full attention is then applied on the concatenation $\texttt{Cat}(\cdot)$ of the extracted features, \ie, $(\texttt{Cat}(Q_{X^{\prime}}, Q_{text}), \texttt{Cat}(K_{X^{\prime}}, K_{text}), \texttt{Cat}(V_{X^{\prime}}, V_{text}))$.
Instead of using the standard full attention, our Swin-MMDiT employs two types of window attention: a regular window attention starting from the top-left unflattened pixel of $X$, and a shifted window attention, offset by half the window size from the regular windows.

As shown in Figure~\ref{fig:network:b}, the first transformer block uses regular window attention with a $t\times h \times w$ window.
Specifically, the video feature $X$ is first divided into $(\frac{T}{t} + 1) \times (\frac{H}{h} + 1) \times (\frac{W}{w} + 1)$ windows, with some windows smaller than $t \times h \times w$.
In Swin Transformer~\cite{liu2021swin,liu2021video}, a cyclic-shifting strategy with a masking mechanism is required to make the window size divisible by the feature map size.
In contrast, our Swin-MMDiT benefits from the flexibility of NaViT and Flash attention~\cite{dao2023flashattention2}. Here, the partitioned window features are flattened into a concatenated 2D tensor, and attention is calculated within each window, eliminating the need for complex masking strategies on the 3D feature map. 
The subsequent transformer block applies shifted window attention, where windows are offset by $(\frac{t}{2}, \frac{h}{2}, \frac{w}{2})$ before attention is calculated similarly to regular window attention. 
For attention calculations, we replace the absolute 2D positional frequency embeddings used in SD3 with 3D relative rotary positional embeddings (RoPE)~\cite{su2024roformer} within each window, avoiding the resolution bias introduced by positional embeddings.

As shown in Figure~\ref{fig:network}, for simplicity, we use separate attention mechanisms for video and text features instead of the single multi-modality attention in MMDiT~\cite{esser2024scaling}.
Specifically, the \textit{key} and \textit{value} of the video window features and text features are concatenated. We then compute attention by calculating the similarity between the concatenated \textit{key} and \textit{value} with the \textit{query} of the video window and text features, respectively.
This approach does not increase computational cost, and in practice, we observe no significant drop in performance.

\begin{figure*}[th]
\begin{center}
    \centering
    \includegraphics[width=\linewidth]{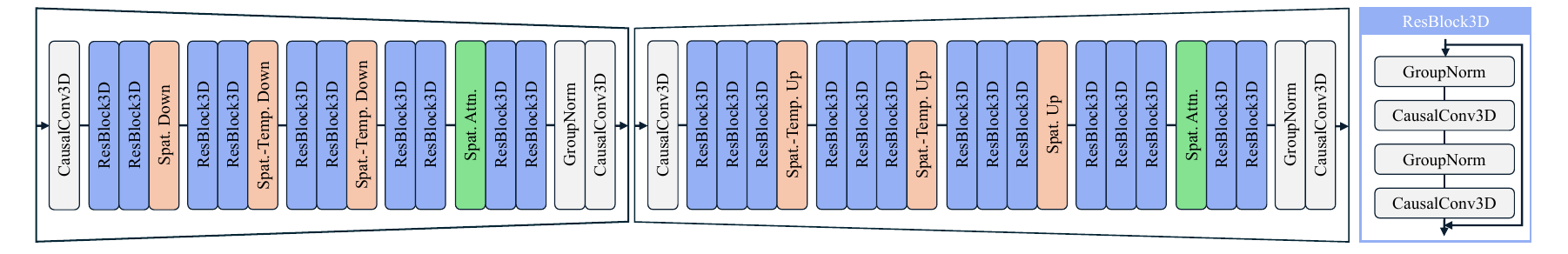}
    \vspace{-5mm}
    \captionof{figure}{
     The model architecture of casual video autoencoder. In contrast to naively inflating an existing image autoenoder, we redesign a casual video VAE with spatial-temporal compression capability to achieve a strong reconstruction capability.
    } \vspace{-10mm}
    \label{fig:network_vae}
\end{center}%
\end{figure*}

\subsection{Causal Video VAE}
\label{subsec:vae}
To process video input, existing diffusion-based VR methods~\cite{zhou2024upscaleavideo,yang2023mgldvsr,he2024venhancer} typically fine-tune a pretrained image autoencoder for video by inserting 3D convolution layers.
Without temporal compression, these video autoencoders are inefficient for both training and inference.
Moreover, the limited number of latent channels (\,  i.e., $4$) prevents these autoencoders from reconstructing videos with high quality.
Instead of fine-tuning a pretrained image autoencoder, we train a video autoencoder from scratch with the following improvements: 1) We use a causal 3D residual block rather than a vanilla 3D block to capture spatial-temporal representations. In this way, our video autoencoder is capable of handling long videos by cutting them into clips. 
2) We increase the latent channels to $16$ following SD3~\cite{esser2024scaling}, to increase the model capacity for better reconstruction.
3) We apply a temporal compression factor of $4$ for more efficient video encoding.
The overall architecture is shown in Figure~\ref{fig:network_vae}.
We follow the common practice~\cite{esser2024scaling} to train our casual video VAE on a large dataset with $\ell 1$ loss, LPIPS loss~\cite{zhang2018perceptual} and GAN loss~\cite{goodfellow2020generative}.

\subsection{Large-scale Training}
\label{subsec:training}
Training a large-scale VR model on millions of high-resolution videos is challenging and remains under-explored.
Existing VR approaches~\cite{zhou2024upscaleavideo,he2024venhancer,yang2023mgldvsr} are trained on limited resources, which inevitably hinders their ability to generalize to more complex, real-world VR tasks.
Besides model architectures, we extend our exploration to include more diverse training data and training strategies to scale up the training. 

\noindent\textbf{Large-scale Mixed Data of Images and Videos.} 
By virtue of the flexibility of our model architecture, we can train the model on image and video data simultaneously. To this end, we first collect a large-scale mixed dataset of images and videos.
Specifically, we collect about $10$ million images and $5$ million videos.
The images vary in resolution, with most exceeding $1024 \times 1024$ pixels.
The videos are $720$p, randomly cropped from higher-resolution videos to improve training efficiency.
In practice, we observe that cropping yields better performance than resizing.
To ensure high-quality data, we further apply several evaluation metrics~\cite{laion_aes,wang2022exploring,wu2022fastquality,ke2021musiq} to filter out low-quality samples.

\noindent\textbf{Precomputing Latents and Text Embeddings.} 
Training on high-resolution data poses challenges for training efficiency due to the slow encoding speed required to convert large videos into latent space with the pretrained VAE.
In practice, encoding a 720p video with $21$ frames takes approximately $2.9$s on average, roughly as long as a single forward pass of the diffusion transformer model.
In addition, encoding the low-quality (LQ) condition also requires VAE processing, doubling the encoding time per training iteration.
By precomputing high-quality (HQ) and LQ video latent features along with text embeddings, we can achieve a $4 \times$ speed up in training.
Applying diverse degradations on large-scale data also ensures sufficient random degradations applied to LQ conditions, which is crucial for training real-world VR models.
Furthermore, eliminating the need to load pretrained VAE and text models saves GPU memory, allowing for a larger batch size for training.

\noindent\textbf{Progressively Growing Up of Resolution and Duration.}
Our model is trained based on SD3-Medium~\cite{esser2024scaling} with 2.2B parameters.
Although SD3-Medium handles high resolutions, \eg, $1024 \times 1024$, we found it challenging to directly adapt it into a VR model with our architecture at that resolution.
Instead, we begin by tuning on short, low-resolution videos ($5$ frames at $256 \times 256$) and progressively increase the lengths and resolutions to ($9$ frames at $512 \times 512$) and eventually $21$ frames at $768 \times 768$.
The final model is trained on data with varying lengths and resolutions.
We observe a rapid convergence with this progressive tuning strategy.

\noindent\textbf{Injecting Noise to Condition.}
We follow existing methods~\cite{wang2021realesrgan,zhou2024upscaleavideo,chan2022investigating} to create synthetic LQ-HQ image and video pairs for training.
While effective, we observe a degradation gap between synthetic LQ videos and real-world ones, as synthetic videos typically exhibit much more severe degradations than those found in real-world videos.
Simply lowering the degradation level for synthetic training data could weaken the model's generative ability, so instead, we follow the strategy of injecting random noise to the latent LQ condition~\cite{zhou2024upscaleavideo,blattmann2023stable}. This is done by diffusing the condition as $C^{\tau}_\text{LQ} = \alpha_{\tau} C_\text{LQ} + \sigma_{\tau} \bm{\epsilon}$, where $\bm{\epsilon} \sim \mathcal{N}(\bm{0}, \bm{I})$, $\tau$ is the noise level associated with the early steps in the noise schedule defined by $\alpha_t$ and $\sigma_t$.

Besides adding noise to the LQ condition, 
we enable the flexible use of the text encoder by randomly replacing the text input to each of the three text encoders with null prompts, similar to SD3~\cite{esser2024scaling}.
Although a similar approach could be applied to LQ conditions to enhance the model's generative capability, we found that excessively strong generative ability often results in reduced output fidelity. Therefore, we opted not to include it in the final model.

%% file: sec/4_exp.tex
\section{Experiments}
\label{sec:exp}

\begin{table*}[th]
    \begin{center}
    \setlength{\fboxsep}{2.1pt}
    \caption{
        Quantitative comparisons on VSR benchmarks from diverse sources, \ie, synthetic (SPMCS, UDM10, REDS30, YouHQ40), real (VideoLQ), and AIGC (AIGC38) data. The best and second performances are marked in \colorbox{rred}{\underline{red}} and \colorbox{oorange}{orange}, respectively.
        }
    \label{tab:comparison}
    \vspace{-2mm}
    \renewcommand{\arraystretch}{1.15}
    \renewcommand{\tabcolsep}{1.8mm}
    \resizebox{\linewidth}{!}{
    \begin{tabular}{l|c|c|c|c||c|c|c|c|c}
    \hline
    Datasets & Metrics & Real-ESRGAN~\cite{wang2021realesrgan} &  SD $\times$4 Upscaler~\cite{sdupscaler} & ResShift~\cite{yue2024efficient} & RealViFormer~\cite{zhang2024realviformer} & MGLD-VSR~\cite{yang2023mgldvsr}  & Upscale-A-Video~\cite{zhou2024upscaleavideo} & VEhancer~\cite{he2024venhancer} & Ours \\
    \hline
    \hline
    \multirow{8}{*}{SPMCS} & PSNR $\uparrow$ & 22.55 & 22.75 & 23.14 & \colorbox{rred}{\underline{24.19}} & \colorbox{oorange}{23.41} & 21.69 & 18.20 & 22.37 \\
         & SSIM $\uparrow$ 	& \colorbox{oorange}{0.637}	& 0.535 & 0.598 & \colorbox{rred}{\underline{0.663}} &	0.633 &	0.519 & 0.507 & 0.607 \\
         & LPIPS $\downarrow$ & 0.406 & 0.554 & 0.547 &	0.378 & \colorbox{oorange}{0.369} & 0.508 & 0.455 & \colorbox{rred}{\underline{0.341}} \\
         & DISTS $\downarrow$ & 0.189 & 0.247 &	0.261 &	0.186 &	\colorbox{oorange}{0.166} & 0.229 & 0.194 &	\colorbox{rred}{\underline{0.141}}  \\
         \cline{2-2}
         & NIQE $\downarrow$ &	3.355	& 5.883	& 6.246 & 3.431	& 3.315 &	\colorbox{oorange}{3.272}	& 4.328	&\colorbox{rred}{\underline{3.207}} \\
         & MUSIQ $\uparrow$ 	& 62.78	& 42.09 & 55.11 &	62.09 &	\colorbox{rred}{\underline{65.25}} &	\colorbox{oorange}{65.01} &	54.94 &	64.28 \\
         & CLIP-IQA $\uparrow$ &	0.451 &	0.402 &	\colorbox{rred}{\underline{0.598}} &	0.424 &	0.495 &	0.507 &	0.334 &	\colorbox{oorange}{0.587} \\
         & DOVER $\uparrow$ & \colorbox{oorange}{8.566} &	4.413 &	5.342 &	7.664 &	8.471 &	6.237 &	7.807 &	\colorbox{rred}{\underline{10.508}} \\
    \hline
    \multirow{8}{*}{UDM10} & PSNR $\uparrow$ &	24.78	& 26.01	& 25.56 &	\colorbox{rred}{\underline{26.70}}	& \colorbox{oorange}{26.11} &	24.62	& 21.48	& 25.76 \\
         & SSIM $\uparrow$ 	& 0.763	& 0.698 &	0.743 &		\colorbox{rred}{\underline{0.796}} &	\colorbox{oorange}{0.772} &	0.712 &	0.691 & 0.771 \\
         & LPIPS $\downarrow$ &	\colorbox{oorange}{0.270} &	0.424 &	0.417 &	0.285 &	0.273 &	0.323 &	0.349 &	\colorbox{rred}{\underline{0.231}} \\
         & DISTS $\downarrow$ &	0.156 &	0.234 &	0.211 &	0.166 & \colorbox{oorange}{0.144} &	0.178 &	0.175 &	\colorbox{rred}{\underline{0.116}} \\
         \cline{2-2}
         & NIQE $\downarrow$ &	4.365	& 6.014	& 5.941 &	3.922	& 3.814 &	\colorbox{rred}{\underline{3.494}}	& 4.883	& \colorbox{oorange}{3.514} \\
         & MUSIQ $\uparrow$ 	& 54.18	& 30.33 &	51.34 &	55.60 &	58.01 & \colorbox{oorange}{58.31} &	46.37 &	\colorbox{rred}{\underline{59.14}} \\
         & CLIP-IQA $\uparrow$ &	0.398 &	0.277 &	\colorbox{rred}{\underline{0.537}} &	0.397 &	0.443 &	0.458 &	0.304 &	\colorbox{oorange}{0.524} \\
         & DOVER $\uparrow$ &	7.958 &	3.169 &	5.111 &	7.259 &	7.717 &	\colorbox{oorange}{9.238} &	8.087 &	\colorbox{rred}{\underline{10.537}} \\
    \hline
    \multirow{8}{*}{REDS30}& PSNR $\uparrow$ &	21.67	& \colorbox{oorange}{22.94}	& 22.72 &	\colorbox{rred}{\underline{23.34}}	& 22.74 &	21.44	& 19.83	& 20.44 \\
         & SSIM $\uparrow$ 	& 0.573	& 0.563 &	0.572 &	\colorbox{rred}{\underline{0.615}} &	\colorbox{oorange}{0.578} &	0.514 &	0.545 &	0.534 \\
         & LPIPS $\downarrow$ &	0.389 &	0.551 &	0.509 &	\colorbox{oorange}{0.328} &	\colorbox{rred}{\underline{0.271}} &	0.397 &	0.508 &	0.346 \\
         & DISTS $\downarrow$ &	0.179 &	0.268 &	0.234 &	0.154 &	\colorbox{rred}{\underline{0.097}} &	0.181 &	0.229 &	\colorbox{oorange}{0.138} \\
         \cline{2-2}
         & NIQE $\downarrow$ &	2.879	& 6.718	& 6.258 &	3.032	& \colorbox{rred}{\underline{2.550}} &	\colorbox{oorange}{2.561}	& 4.615	& 2.729 \\
         & MUSIQ $\uparrow$ 	& 57.97	& 25.57 & 47.50 &	\colorbox{oorange}{58.60} &	\colorbox{rred}{\underline{62.28}} &	56.39 &	37.95 &	57.55 \\
         & CLIP-IQA $\uparrow$ &	0.403 &	0.202 &	\colorbox{rred}{\underline{0.554}} &	0.392 &	0.444 &	0.398 &	0.245 &	\colorbox{oorange}{0.451} \\
         & DOVER $\uparrow$ & 5.552 &	2.737 &	3.712 &	5.229 &	\colorbox{oorange}{6.544} &	5.234 &	5.549 &	\colorbox{rred}{\underline{6.673}} \\
    \hline
    \multirow{8}{*}{YouHQ40}& PSNR $\uparrow$ &	22.31	& 22.51	& \colorbox{oorange}{22.67} &	\colorbox{rred}{\underline{23.26}}	& 22.62 &	21.32	& 18.68	& 21.15 \\
         & SSIM $\uparrow$ 	& \colorbox{oorange}{0.605}	& 0.528 &	0.579 &	\colorbox{rred}{\underline{0.606}} &	0.576 &	0.503 &	0.510 &	0.554 \\
         & LPIPS $\downarrow$ &	\colorbox{oorange}{0.342} &	0.518 & 0.432 &	0.362 &	0.356 &	0.404 &	0.449 &	\colorbox{rred}{\underline{0.298}} \\
         & DISTS $\downarrow$ &	0.169 &	0.242 &	0.215 &	0.193 &	\colorbox{oorange}{0.166} &	0.196 &	0.175 &	\colorbox{rred}{\underline{0.118}} \\
         \cline{2-2}
         & NIQE $\downarrow$ &	3.721	& 5.954	& 5.458 &	3.172	& 3.255 &	\colorbox{oorange}{3.000}	& 4.161	&\colorbox{rred}{\underline{2.913}} \\
         & MUSIQ $\uparrow$ 	& 56.45	& 36.74 &	54.96 &	61.88 &	63.95 &	\colorbox{oorange}{64.450} &	54.18 &	\colorbox{rred}{\underline{67.45}} \\
         & CLIP-IQA $\uparrow$ &	0.371 &	0.328 &	\colorbox{oorange}{0.590} &	0.438 &	0.509 &	0.471 &	0.352 &	\colorbox{rred}{\underline{0.635}} \\
         & DOVER $\uparrow$ &	10.92 &	5.761 & 7.618 &	9.483 &	10.503 &	9.957 &	\colorbox{oorange}{11.444} &	\colorbox{rred}{\underline{12.788}} \\
    \hline
    \hline
    \multirow{4}{*}{VideoLQ} & NIQE $\downarrow$&	4.014	& 4.584 &	4.829 & 4.007 &	3.888 &	\colorbox{rred}{3.490} &	4.264	&	\colorbox{oorange}{\underline{3.874}}\\
         & MUSIQ $\uparrow$	& \colorbox{rred}{\underline{60.45}}	& 43.64	& \colorbox{oorange}{59.69}	& 57.50 &	59.50	& 58.31 &	52.59		& 54.41 \\
         & CLIP-IQA $\uparrow$&	0.361	& 0.296 &	\colorbox{rred}{\underline{0.487}} &	0.312 &	0.350 &	\colorbox{oorange}{0.371} &	0.289	&	0.355 \\
         & DOVER $\uparrow$ 	& \colorbox{oorange}{8.561} & 4.349 &	6.749 &	6.823 &	7.325	& 7.090 &	\colorbox{rred}{\underline{8.719}}	&	8.009 \\
    \hline
    \multirow{4}{*}{AIGC38} & NIQE $\downarrow$ &	4.942	& 4.399	& 4.853 &	4.444 &	4.162 &	\colorbox{oorange}{4.124}	& 4.759	&	\colorbox{rred}{\underline{3.955}} \\
         & MUSIQ $\uparrow$ &	58.39	& 56.72 &	\colorbox{oorange}{64.38}	& 58.73 & 62.03 &	63.15 &	53.36	&	\colorbox{rred}{\underline{65.91}} \\
         & CLIP-IQA $\uparrow$&	0.442	& 0.554 &	\colorbox{rred}{\underline{0.660}} &	0.473 &	0.528 &	0.497 &	0.395	&	\colorbox{oorange}{0.638} \\
         & DOVER $\uparrow$ & 12.275 & 10.547	& 12.082	& 10.245 &	11.008 &	\colorbox{oorange}{12.857}	& 12.178	&	\colorbox{rred}{\underline{13.424}} \\
    \hline
    \end{tabular}
    }
    \end{center}
    \vspace{-8mm}
\end{table*}

\noindent\textbf{Implementation Details.}
We train SeedVR on 256 NVIDIA H100-80G GPUs with around 150 720p frames per batch per GPU.
We initialize the model parameters from SD3-Medium~\cite{esser2024scaling} and train the full model following the strategies discussed in Sec.~\ref{subsec:training}.
We mostly follow the training settings in SD3~\cite{esser2024scaling} to train the diffusion transformer.
We follow Upscale-A-Video~\cite{zhou2024upscaleavideo} to synthesize training pairs. 
The entire training process requires about 30K H100-80G GPU hours.
As for the training of our casual video VAE, we follow the standard settings in SD3~\cite{esser2024scaling} and train on internal data with a resolution of $17 \times 256 \times 256$.
The model is trained on 32 NVIDIA H100-80G GPUs with a batch size of $5$ per GPU for 115,000 iterations.

\noindent\textbf{Experimental Settings.}
We use various metrics to evaluate both the frame quality and overall video quality.
For synthetic datasets with LQ-HQ pairs, we employ full-reference metrics such as PSNR, SSIM, LPIPS~\cite{zhang2018unreasonable}, and DISTS~\cite{ding2020image}, and no-reference metrics including NIQE~\cite{mittal2012making}, CLIP-IQA~\cite{wang2022exploring}, MUSIQ~\cite{ke2021musiq}, and DOVER~\cite{wu2023exploring}.
For real-world and AIGC test data, we adopt no-reference metrics, \ie, NIQE, CLIP-IQA, MUSIQ, and DOVER due to the absence of ground truth.
To ensure fair comparisons, all testing videos are processed to be 720p while maintaining the original length.
We follow previous work~\cite{zhou2024upscaleavideo} to test on synthetic benchmarks \ie, SPMCS~\cite{PFNL}, UDM10~\cite{tao2017spmc}, REDS30~\cite{nah2019ntire}, and YouHQ40~\cite{zhou2024upscaleavideo}, using the same degradations as training.
Additionally, we evaluate the models on a real-world dataset (\ie, VideoLQ~\cite{chan2022investigating}) and an AIGC dataset (\ie, AIGC38) that collects 38 AI-generated videos.

\begin{figure*}[th]
\begin{center}
    \centering
    \includegraphics[width=\linewidth]{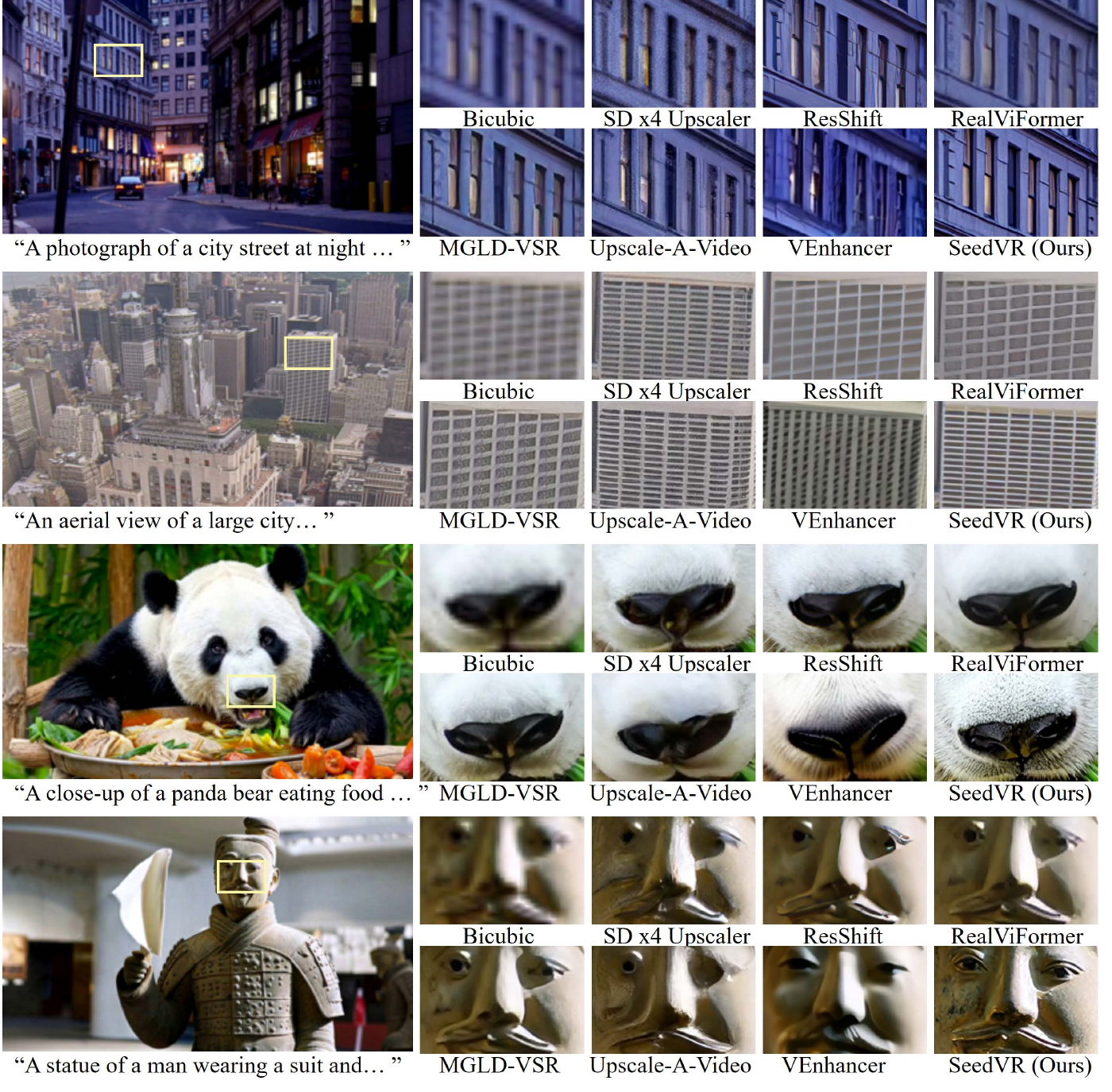}
    \vspace{-5mm}
    \captionof{figure}{
     Qualitative comparisons on both real-world videos in VideoLQ~\cite{chan2022investigating} dataset (the first and second row) and AIGC dataset (the third and fourth row). Our SeedVR is capable of generating realistic details.
    When compared to existing methods, it notably excels in its restoration capabilities, successfully removing the degradations while maintaining the textures of the buildings, the panda's nose and the face of the terracotta warrior.
    \textbf{(Zoom-in for best view)}.
    } \vspace{-8mm}
    \label{fig:compare}
\end{center}%
\end{figure*}

\begin{table*}[ht]
\setlength{\tabcolsep}{4pt}
\begin{center}
\setlength{\fboxsep}{2.0pt}
\caption{Quantitative comparisons on VAE models commonly used in existing latent diffusion models~\cite{rombach2021highresolution,he2024venhancer,cosmos,opensora,pku_yuan_lab_and_tuzhan_ai_etc_2024_10948109,yang2024cogvideox}. The best and second performances are marked in \colorbox{rred}{\underline{red}} and \colorbox{oorange}{orange}, respectively.}
\label{tab:vae_main}
\vspace{-2mm}
\renewcommand{\arraystretch}{1.11}
\renewcommand{\tabcolsep}{1.8mm}
\scalebox{0.9}{
\begin{tabular}{l|c|c|c|c|c|c|c|c}
    \hline
    \makecell{Methods \\ (VAE)} & \makecell{Params \\ (M)} & \makecell{Temporal \\ Compression} & \makecell{Spatial \\ Compression} & \makecell{Latent \\ Channel} & PSNR $\uparrow$ & SSIM $\uparrow$ & LPIPS $\downarrow$ & rFVD $\downarrow$\\
    \hline
    \hline
    SD 2.1~\cite{rombach2021highresolution} & 83.7 & - & 8 & 4 & 29.50 & 0.9050 & 0.0998 & 8.14 \\
    VEnhancer~\cite{he2024venhancer} & 97.7 & - & 8 & 4 & 30.81 & 0.9356 & 0.0751 & 11.10  \\
    Cosmos~\cite{cosmos} & 90.2 & 4 & 8 & 16 & 32.34 & 0.9484 & 0.0847 & 13.02 \\
    OpenSora~\cite{opensora} & 393.3 & 4 & 8 & 4 & 27.70 & 0.8893 & 0.1661 & 47.04  \\
    OpenSoraPlan v1.3~\cite{pku_yuan_lab_and_tuzhan_ai_etc_2024_10948109} & 147.3 & 4 & 8 & 16 & 30.41 & 0.9280 & 0.0976 & 27.70\\
    CV-VAE (SD3)~\cite{zhao2024cv} & 181.9 & 4 & 8 & 16 & 33.21 & 0.9612 &  \colorbox{oorange}{0.0589} & 6.50 \\
    CogVideoX~\cite{yang2024cogvideox} & 215.6 & 4 & 8 & 16 & \colorbox{rred}{\underline{34.30}} & \colorbox{rred}{\underline{0.9650}} & 0.0623 & \colorbox{oorange}{6.06}  \\
    \hline
    Ours   & 250.6 & 4 & 8 & 16 & \colorbox{oorange}{33.83} & \colorbox{oorange}{0.9643} & \colorbox{rred}{\underline{0.0517}} & \colorbox{rred}{\underline{1.85}} \\
    \hline
\end{tabular}
}
\end{center}
\vspace{-8mm}
\end{table*}

\subsection{Comparison with Existing Methods}
\noindent\textbf{Quantitative Comparisons.}
As shown in Table \ref{tab:comparison}, our method achieves significantly superior performance on 4 out of 6 benchmarks (\ie, SPMCS, UDM10, YouHQ40, and AIGC38). Similar to other diffusion-based methods, our SeedVR shows limitations on certain metrics like PSNR and SSIM~\cite{yu2024scaling,wang2024exploiting,yue2022difface} on these benchmarks because these metrics are primarily designed to measure pixel-level fidelity and structural similarity, while SeedVR focuses on perceptual quality.
As can be observed, SeedVR obtains high DISTS, LPIPS, NIQE and DOVER scores, indicating the high perceptual quality of its generated results.
It is worth noting that MGLD-VSR and RealViFormer are trained on REDS~\cite{nah2019ntire}, which explains their strong performance on the corresponding test set, REDS30.
Even so, our approach remains competitive, achieving the best DOVER score on REDS30.
The consistent superiority across datasets from various sources demonstrates the effectiveness of our method.

\noindent\textbf{Qualitative Comparisons.}
Figure~\ref{fig:compare} shows visual results on both real-world~\cite{chan2022investigating} and AIGC videos.
Our seedVR outperforms existing VR approaches by a large margin in both degradation removal and texture generation.
Specifically, SeedVR effectively recovers detailed structures, such as the building architectures with severely degraded video inputs.
For AIGC videos, SeedVR faithfully restores fine details, such as the panda's nose and the terracotta warrior's face in Figure~\ref{fig:compare}, where other approaches produce blurred details.

\subsection{Ablation Study}
\noindent\textbf{Effectiveness of Casual Video VAE.}
We first examine the significance of the proposed casual video VAE.
As shown in Table \ref{tab:vae_main}, our VAE demonstrates better video reconstruction quality.
Compared to state-of-the-art VAE models for video generation and restoration, our VAE reaches the lowest rFVD~\cite{unterthiner2018towards} score, 69.5\% lower than the second best.
Besides, our VAE achieves the best LPIPS score and competitive PSNR and SSIM relative to CogVideoX~\cite{yang2024cogvideox}, indicating its superior reconstruction capability.

\noindent\textbf{Window Size for Attention.}
Besides the powerful VAE, a key aspect of our design is the flexible window attention, which enables restoration at arbitrary resolutions.
We measure the performance with different window sizes.
Specifically, we train the model using different window sizes under the same settings for 12.5k iterations.
Results show that smaller window sizes significantly increase training time.
As shown in Table~\ref{tab:window_time}, training time rises considerably with smaller windows;
for instance, with $1\times8\times8$ window, the training time required is $455.49$, which is 19.24 times longer than a $1\times64\times64$ window.
This increase is due to each window being assigned a text prompt in the attention computation, introducing text guidance while retaining flexibility for arbitrary resolutions. Therefore, using larger window sizes reduces the number of text tokens required for attention, improving both training and inference efficiency.

We further validate the performance under different window sizes on the YouHQ40 dataset~\cite{zhou2024upscaleavideo} measured by DOVER~\cite{wu2023exploring}.
From Table~\ref{tab:window_dover}, we make the following observations: 1) The performance of full spatial attention declines as the temporal window length increases. We believe this is due to the high token count in full attention, which requires a much longer training period, far beyond 12.5k iterations, to converge fully. Larger temporal windows amplify this need. 
2) Smaller spatial windows, \eg, $32 \times 32$, outperform full attention, but still show a performance drop as temporal length increases. We hypothesize that smaller temporal and spatial windows, like \eg, $1\times32\times32$, allow for faster convergence, leading to better performance. However, smaller spatial windows may face difficulties in capturing temporal dependencies, requiring additional training to prevent performance degradation.
3) For a spatial window size of $64 \times 64$, performance is comparable with shorter temporal lengths, \ie, $1$ and $3$.
Increasing the window length to 5 notably improves results, likely because the larger window size captures long-range dependencies and enhances semantic alignment between text prompts and restoration.
These observations validate our design choice of using a $5 \times 64 \times 64$ attention window.

\begin{table}[t]
\setlength{\tabcolsep}{4pt}
\begin{center}
\setlength{\fboxsep}{2.1pt}
\caption{Training efficiency (sec/iter) with different window sizes. }
\label{tab:window_time}
\vspace{-1mm}
\renewcommand{\arraystretch}{1.11}
\renewcommand{\tabcolsep}{1.8mm}
\resizebox{0.40\textwidth}{!}{
\begin{tabular}{c|c|c|c|c}
    \hline
    \multirow{2}{*}{\makecell[c]{Temp. Win. \\ Length}} & \multicolumn{4}{c}{Spat. Win. Size} \\
    \cline{2-5} & $8 \times 8$ & $16 \times 16$ & $32 \times 32$ & $64 \times 64$ \\
    \hline
    $t=1$ & 455.49 & 138.29 & 58.37 & 23.68 \\
    $t=5$ & 345.78 & 110.01 & 46.49 & 20.29 \\
    \hline
\end{tabular}
}
\end{center}
\vspace{-5mm}
\end{table}

\begin{table}[t]
\setlength{\tabcolsep}{4pt}
\begin{center}
\setlength{\fboxsep}{2.1pt}
\caption{Ablation study on the performance of different window sizes. All baselines are trained on 16 NVIDIA A100-80G cards for 12.5k iterations. The comparison is conducted on YouHQ40~\cite{zhou2024upscaleavideo} synthetic data and measured by DOVER ($\uparrow$)~\cite{wu2023exploring}.}.
\label{tab:window_dover}
\vspace{-2mm}
\renewcommand{\arraystretch}{1.15}
\renewcommand{\tabcolsep}{1.8mm}
\resizebox{0.33\textwidth}{!}{
\begin{tabular}{c|c|c|c}
    \hline
    \multirow{2}{*}{\makecell[c]{Temp. Win. \\ Length}} & \multicolumn{3}{c}{Spat. Win. Size} \\
    \cline{2-4} & $32 \times 32$ & $64 \times 64$ & Full \\
    \hline
    $t=1$ & 11.947 & 10.690 & 10.799 \\
    $t=3$ & 11.476 & 10.429 & 9.145 \\
    $t=5$ & 10.558 & 11.595 & 8.521 \\
    \hline
\end{tabular}
}
\end{center}
\vspace{-7mm}
\end{table}

%% file: sec/5_conclusion.tex
\section{Conclusion}
\label{sec:conclusion}
We have presented SeedVR, a novel diffusion transformer model designed as foundational architecture to tackle high-quality VR with arbitrary resolutions and lengths. SeedVR builds on the strengths of existing diffusion-based methods, yet addresses key limitations through a flexible architecture that combines a large attention window, a causal video autoencoder, and efficient training strategies. Extensive experiments demonstrate SeedVR's superior ability to handle both synthetic and real-world degradations, with improved visual realism and detail consistency across frames. Notably, SeedVR is over twice as fast as existing methods despite its larger parameter size. In the future, we will improve the sampling efficiency and robustness of SeedVR.